\documentclass[runningheads]{llncs}
  
\usepackage{graphicx}
\usepackage{multirow}
\usepackage{eucal}
\usepackage{multicol}
\usepackage{booktabs}
\usepackage{subfig}
\usepackage{amsmath}
\usepackage{amsmath,amssymb,amsfonts}
\usepackage{algorithm}
\usepackage{textcomp}
\usepackage[table]{xcolor}
\definecolor{lightgray}{gray}{0.9}
\usepackage{booktabs}
\usepackage{lineno}
\usepackage[table]{xcolor}%
\usepackage{xcolor,soul}
\sethlcolor{lightgray}
\usepackage{gensymb}
\usepackage{amssymb}
\usepackage{color}
\usepackage[colorlinks=true,
            linkcolor=red,
            citecolor=blue]{hyperref}
\usepackage{indentfirst}
\usepackage[noend]{algpseudocode}
\def\algbackskip{\hskip-\ALG@thistlm}
\usepackage{caption}
\usepackage{helvet}
\usepackage{courier}
\usepackage{bm}
\usepackage{epsfig}
\usepackage{enumitem}
\usepackage{rotating}
\usepackage[T1]{fontenc}
\usepackage{booktabs}
\usepackage{subfig}
\usepackage{url,longtable,multirow,morefloats,stfloats, floatflt,cancel,tfrupee}
\makeatletter
\AtBeginDocument{\@ifpackageloaded{textcomp}{}{\usepackage{textcomp}}}
\makeatother
\usepackage{colortbl,xcolor}
\usepackage{pifont}
\usepackage[nointegrals]{wasysym}
\usepackage[format=plain,
            labelfont=it,
            textfont=it]{caption}
\makeatletter
\def\algbackskip{\hskip-\ALG@thistlm}

\@ifundefined{etal}{}{}

\usepackage{ifxetex}
\ifxetex\else\if@twocolumn\@ifpackageloaded{stfloats}{}{\usepackage{dblfloatfix}}\fi\fi

\graphicspath{{fig/}}
\begin{document}
\title{Imbalance-Aware Self-Supervised Learning for \\ 3D Radiomic Representations}
\author{Hongwei Li\inst{1,2} \and
    Fei-Fei Xue\inst{4} \and
    Krishna Chaitanya \inst{3} \and
    Shengda~Luo\inst{5} \and
    Ivan~Ezhov\inst{1} \and
   Benedikt~Wiestler\inst{6} \and 
  Jianguo~Zhang\inst{4} \and
  Bjoern~Menze\inst{1,2} }
  

\institute{Department of Computer Science, Technical University of Munich, Germany \and
Department of Quantitative Biomedicine, University of Zurich, Switzerland \and
ETH Zurich, Switzerland \and
 Department of Computer Science and Engineering, Southern University of Science and Technology, Shenzhen, China \and 
 Faculty of Information Technology, Macau University of Science and Technology \and 
 Klinikum rechts der Isar, Technical University of Munich, Germany \\
\email{Email: hongwei.li@tum.de}}
\maketitle              %
\begin{abstract}
Radiomics can quantify the properties of regions of interest in medical image data. Classically, they account for pre-defined statistics of shape, texture, and other low-level image features. Alternatively, deep learning-based representations are derived from supervised learning but require expensive annotations and often suffer from overfitting and data imbalance issues.
In this work, we address the challenge of learning the representation of a 3D medical image for an effective quantification under data imbalance. We propose a \emph{self-supervised} representation learning framework to learn high-level features of 3D volumes as a complement to existing radiomics features.
Specifically, we demonstrate how to learn image representations in a self-supervised fashion using a 3D Siamese network. More importantly, we deal with data imbalance by exploiting two unsupervised strategies: a) sample re-weighting, and b) balancing the composition of training batches. 
When combining the learned self-supervised feature with traditional radiomics, we show significant improvement in brain tumor classification and lung cancer staging tasks covering MRI and CT imaging modalities. Codes are available in \url{https://github.com/hongweilibran/imbalanced-SSL}. 
 
\end{abstract}

\section{Introduction}
Great advances have been achieved in supervised deep learning, reaching expert-level performance on some considerably challenging applications \cite{esteva2017dermatologist}. However, supervised methods for image classification commonly require relatively large-scale datasets with ground-truth labels which is time- and resource-consuming in the medical field. 
Radiomics is a translational field aiming to extract objective and quantitative information from clinical imaging data. While traditional radiomics methods, that rely on statistics of shape, texture and others \cite{aerts2014decoding}, are proven to be generalizable in various tasks and domains, their discriminativeness is often not guaranteed since they are low-level features which are not specifically optimized on target datasets. 

Self-supervised learning for performing \emph{pre-text tasks} have been explored in medical imaging \cite{zhou2019models,zhuang2019self}, that serve as a proxy task to pre-train the deep neural networks. They learn representations commonly in a supervised manner on proxy tasks. Such methods depend on heuristics to design pre-text tasks which could limit the discriminativeness of the learnt representations. In this work, we investigate self-supervised \emph{representation learning} which aims to \textbf{directly} learn the representation of the data without a proxy task.
%

Recent contrastive learning-based methods \cite{chen2020simple,he2020momentum,oord2018representation} learn informative representations \emph{without} human supervision. 
However, they often rely on large batches to train and most of them work for 2D images. 
To this end, due to the high dimensionality and limited number of training samples in medical field, applying contrastive learning-based methods may not be practically feasible in 3D datasets. 
Specially, in this study, we identify two main differences required to adapt self-supervised representation learning for radiomics compared to natural image domain: i) 
Medical datasets are often multi-modal and three dimensional. Thus, learning representation methods in {3D} medical imaging would be computationally expensive. ii) heterogeneous medical datasets are \emph{inherently imbalanced}, e.g. distribution disparity of disease phenotypes. Existing methods are built upon approximately balanced datasets (e.g. CIFAR \cite{krizhevsky2009learning} and ImageNet \cite{deng2009imagenet}) and do not assume the existence of data imbalance. 
Thus, how to handle data imbalance problem is yet less explored in the context of self-supervised learning.


\subsubsection{Related work.}
Radiomic 
features have drawn considerable attention due to its predictive power for treatment outcomes and cancer genetics in personalized medicine \cite{gillies2016radiomics,yip2016applications}. Traditional radiomics include shape features,  first-, second-, and higher- order statistics features.

Self-supervised representation learning \cite{bachman2019learning,chen2020simple,chen2020improved,grill2020bootstrap,he2020momentum,ye2019unsupervised} have shown steady improvements with impressive results on multiple natural image tasks, mostly based on contrastive learning \cite{hadsell2006dimensionality}. Contrastive learning aims to attract positive (or \emph{similar}) sample pairs and rebuff negative (or \emph{disimilar}) sample pairs. Positive sample pairs can be obtained by generating two augmented views of one sample, and the remaining samples in the batch can be used to construct the negative samples/pairs for a given positive pair.
In practice, contrastive learning methods benefit from a large number of negative samples. 
In medical imaging, there are some existing work related to contrastive learning \cite{kiyasseh2020clocs,azizi2021big,vu2021medaug}.  
Chaitanya \emph{et al.}'s work \cite{chaitanya2020contrastive} is the most relevant to our study, which proposed a representation learning framework for image segmentation by exploring local and global similarities and dissimilarities. Though these methods are effective in learning representations, they require a large batch size and/or negative pairs, which make them difficult to apply to \emph{3D} medical data. 
Chen \emph{et al.} \cite{chen2020exploring} demonstrates that a Siamese network can avoid the above issues on a 2D network. The Siamese network, which contains two encoders with shared weights, compares two similar representations of two augmented samples from one sample. Importantly, it neither uses negative pairs nor a large batch size. Considering these benefits, we borrow the Siamese structure and extend it to 3D \emph{imbalanced} medical datasets.
\vspace{-0.4cm}
\subsubsection{Contributions.} Our contribution is threefold: (1) We develop a 3D Siamese network to learn self-supervised representation which is high-level and discriminative. (2) For the first time, we explore how to tackle the data imbalance problem in self-supervised learning without using labels and propose two effective unsupervised strategies. (3) We demonstrate that self-supervised representations can complement the existing radiomics and the combination of them outperforms supervised learning in two applications. 

\section{Methodology}
\begin{figure}
    \centering
    \includegraphics[width=0.95\textwidth]{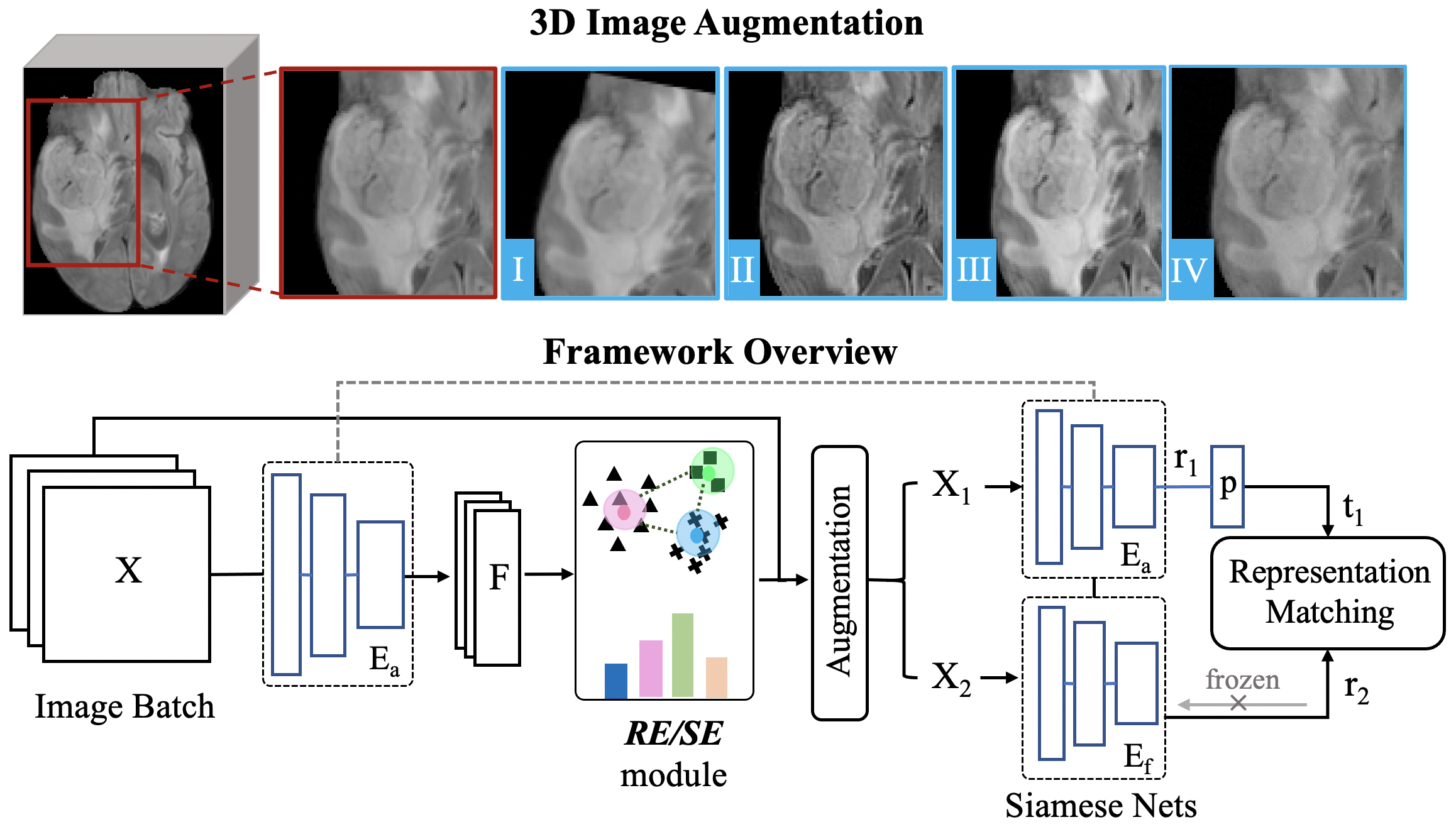}
    \caption{\small Our proposed framework learns invariance from extensive 3D image augmentation within four categories: I) affine transform, II) appearance enhancement, III) contrast change, and IV) adding random noise.  First, an image batch \emph{X} is fed into an initialized 3D encoder to obtain its representation \emph{F}. The \emph{RE/SE} module first estimates its distribution by k-means based clustering and uses two strategies including sample re-weighting (RE) or sample selection (SE) to alleviate data imbalance issue. Each image is randomly augmented into two positive samples \emph{\{X$_{1}$, X$_{2}$\}}  which are then used to train a 3D Siamese network by comparing their representations from two encoders \emph{\{E$_{a}$, E$_{f}$\}} with shared weights. $p$ is a two-layer perceptron to transform the feature.}
    \label{fig:3D_SimSiam}
    \vspace{-0.2cm}
\end{figure} 
The problem of interest is how to learn high-level, discriminative representations on 3D imbalanced medical image datasets in a self-supervised manner. 
The schematic view of the framework is illustrated in Fig. \ref{fig:3D_SimSiam}. First, a pre-trained 3D encoder network, denoted as $E_{a}$, takes a batch of original images ${X}$ with batch size $N$ as input and outputs $N$ representation vectors. The details of the 3D encoder is shown in Tab.~4 of the Supplementary. 
The features are fed into the \emph{RE/SE} module to estimate their individual weights or to resample the batch.

Then each image $x$ in the batch $X$ is randomly augmented into two images (or called an image pair).  They are processed and compared by a 3D Siamese network, nicknamed \emph{3DSiam}, which enjoys relatively low memory without relying on large training batch of 3D data.
The proposed \emph{3DSiam} extends original 2D Siamese network \cite{chen2020exploring} from processing 2D images to 3D volumes while inherits its advantages. 
Since medical datasets are inherently imbalanced, by intuition sole \emph{3DSiam} would probably suffer from imbalanced data distribution. In the following, we first introduce \emph{RE/SE} module to mitigate this issue. 



\subsubsection{RE/SE Module to Handle Imbalance.} Since there is no prior knowledge on the data distribution available, the way to handle imbalance must be \emph{unsupervised}. The vectors mentioned above are fed into a \emph{RE/SE} module before training the \emph{3DSiam} network. The \emph{k}-means algorithm is used first to cluster the representation vectors into $k$ centers. We then proposed two simple yet effective strategies: a) sample re-weighting (RE), and b) sample selection (SE): 

a) Sample re-weighting (RE). Denote a batch with $N$ samples as ${X} = \{x_i|i = 1, 2, ..., N\}$.
Given $k$ clusters, denote the distribution of $k$ clusters of features as ${F} = \{f_j|j = 1, 2, ..., k\}$ over $N$ samples. $f_j$ denotes the frequency of cluster $j$. Then we assign different weights to the samples in each cluster $j$. For each sample $x_i$, representation vector of which belongs to cluster $j$, we assign it a weight of N/$f_j$ to penalize the imbalanced distribution during the batch training. In practice, we further normalize it by re-scaling to guarantee the minimum weight is 1 for each input batch. 

b) Sample selection (SE). Denoting the clusters' centroids as ${C} = \{c_1, c_2, ... c_k\}$, we find the maximum Euclidean distance $max_{i,j \in [1,k], i \neq j}{d(c_i, c_j)}$ among all pairs of centroids. $k$ is a hyper-parameter here. 
We hypothesize that the clusters with maximum centroid distance are representation vectors from different groups. To select $m$ samples from the original $N$ samples to form a new batch, denoted by ${B_c} = \{x_1, x_2, ..., x_m\}$, we sample $\frac{m}{2}$ nearest sample points centered on each of the selected maximum-distance centroids. 
${m}$ is set to be smaller than $\frac{N}{k}$ for low computation complexity and efficient sampling.
The selected new batch is then used to train our \emph{3DSiam} network. A motivation behind the selection strategy is outlined in Supplementary. 

\subsubsection{3D Siamese Network.} The \emph{3DSiam} takes as input two randomly augmented views $x_1$ and $x_2$ from a sample $x$. The two views are processed by two 3D encoder networks with \emph{shared} weights. One of the encoder has frozen weights when training (denoted as $E_{f}$) and the other one is with active weights (denoted as $E_{a}$). 
Before training, $E_{f}$ is always updated to the weights of $E_{a}$. 
$E_{a}$ is followed by a two-layer perceptron called \emph{predictor} to transform the features. 
The final objective is to optimize a matching score between the two \emph{similar} representations $t_{1} \triangleq p\left(E_{a}{(x_{1})}\right)$ and $r_{2} \triangleq E_{f}\left(x_{2}\right)$. \emph{3DSiam} minimizes their negative cosine similarity, which is formulated as:
\begin{equation}
\label{negative_cos}
{S}\left(t_{1}, r_{2}\right)=-\frac{t_{1}}{\left\|t_{1}\right\|_{2}} \cdot \frac{r_{2}}{\left\|r_{2}\right\|_{2}},
\end{equation}
where $\|\cdot\|_{2}$ is $L_{2}$-norm. Following \cite{chen2020simple}, we define a symmetrized loss to train the two encoders, formulated as:  
\begin{equation}
\label{train_loss}
\mathcal{L}=\frac{1}{2} {S}\left(t_{1}, r_{2}\right)+\frac{1}{2} {S}\left(t_{2}, r_{1}\right),
\end{equation}
where $t_{2} \triangleq p\left(E_{f}{(x_{2})}\right)$, $r_{1} \triangleq E_{f}{(x_{1})}$. This loss is defined and computed for each sample with re-weighting in the batch $X$ or the new batch ${B_c}$ with equal weights. Notably the encoder $E_{f}$ on $x_2$ receives no gradient from $r_2$ in the ﬁrst term of Eq. (2), but it receives gradients from $t_2$ in the second term (and vice versa for $x_1$). This training strategy avoids collapsing solutions, i.e., $t_{1}$ and $r_{2}$ outputs a constant over the training process. When training is finished, $r_{2}$ is used as the final representation. 
\section{Experiments}
\subsubsection{Datasets and preprocessing.} The evaluation of our approach is performed on two public datasets: 1) a multi-center MRI dataset (\emph{BraTS}) \cite{menze2014multimodal,bakas2018identifying} including 326 patients with brain tumor. The MRI modalities include FLAIR, T1, T2 and T1-c with a uniform voxel size 1$\times$1$\times$1 \emph{mm$^3$}. Only FLAIR is used in our experiment for simplicity of comparisons. 2) a lung CT dataset with 420 non-small cell lung cancer patients (\emph{NSCLC-radiomics})  \cite{aerts2014decoding,clark2013cancer} \footnote{Two patients were excluded as the ground truth labels are not available.}. The effectiveness of the learnt representations is evaluated on two classification tasks (also called 'down-stream task'): a) discriminating high grade (H-grade) and low grade tumor (L-grade), and b) predicting lung cancer stages (i.e. I, II or III).
The \emph{BraTS} dataset is imbalanced in two aspects: a) the distribution of ground truth labels (H-grade \emph{vs.} L-grade); b) the distribution of available scans among different medical centers. For \emph{NSCLC-radiomics}, the distribution of ground truth labels are imbalanced as well, with ratio of 2:1:6 for stage I, II and III respectively. For \emph{BraTS}, we make use of the segmentation mask to get the centroid and generate a 3D bounding box of 96$\times$96$\times$96 to localize the tumour. If the bounding box exceeds the original volume, the out-of-box region was padded with background intensity. For \emph{NSCLC-radiomics}, we get the lung mask by a recent public lung segmentation tool \cite{hofmanninger2020automatic}, and then generate a 224$\times$224$\times$224 bounding box to localize the lung. The lung volume was then resized to 112$\times$112$\times$112 due to memory constraint. The intensity range of all image volumes was rescaled to [0, 255] to guarantee the success of intensity-based image transformations. 

\subsubsection{Configuration of the training schedule.}
We build a 3D convolutional neural network with two bottleneck blocks as the encoder for all experiments (details in Supplementary). 
In the beginning, we pre-train \emph{3DSiam} for one epoch with a batch size of 6. Then we use it to extract features for the \emph{RE/SE} module. After first epoch, the encoder from the \emph{last iteration} is employed for \emph{dynamic} feature extraction.
For down-stream task evaluations, we use the last-layer feature of the encoder. 
For 3D data augmentation, we apply four categories shown in Fig. \ref{fig:3D_SimSiam}, including random rotations in $[-20, 20]$ degrees, random scale between $[0.7, 1.3]$, and random shift between $[-0.05, 0.05]$, Gamma contrast adjustment between [0.7, 1.5], image sharpening, and Gaussian blurring, considering the special trait of medical images.  For optimization, we use Adam with $10^{-2}$ learning rate and $10^{-4}$ weight decay. Each experiment is conducted using one Nvidia RTX 6000 GPU with 24GB memory. The number of cluster $k$ is set to 3 in all experiments. Its effect is analyzed in the last section.
\vspace{-0.35cm}
\paragraph{Computation Complexity.} For \emph{3DSiam} without \emph{RE/SE} module, the training takes only around four hours for 50 epochs for the results reported for brain tumor classification task. We do not observe significant improvement when increasing the number of epochs after 50. We train \emph{3DSiam} with \emph{RE/SE} module for around 2000 iterations (not epochs) to guarantee that similar number of training images for the models of comparison are involved. 
In \emph{RE/SE} module, the main computation cost is from \emph{k}-means algorithm. We have observed that the overall computation time has increased by 20\% (with i5-5800K CPU). It is worth noting that \emph{RE/SE} module is not required during the inference stage, thus there is no increase of the computational cost in testing. 
\subsubsection{Feature extraction and aggregation.} For each volume, we extract a set of 107 traditional radiomics features \footnote{\url{https://github.com/Radiomics/pyradiomics}} including first- and second-order statistics, shape-based features and gray level co-occurrence matrix, denoted as $f_{trad}$. For the self-supervised learning one, we extract 256 features from the last fully connected layer of the encoder $E_{a}$, denoted as $f_{SSL}$. To directly evaluate the effectiveness of SSL-based features, we concatenate them to a new feature vector $f = [f_{trad}, f_{SSL}]$. Note that $f_{trad}$ and $f_{SSL}$ are always from the same subjects.

\subsubsection{Evaluation protocol, classifier and metrics.} 
For evaluation, we follow the common protocol to evaluate the quality of the pre-trained representations by training a \emph{supervised} linear support vector machine (SVM) classifier on the training set, and then evaluating it on the test set. For binary classification task (BraTS), we use the sensitivity and specificity as the evaluation metrics. For multi-class classification task (lung cancer staging), we report the overall accuracy and minor-class (i.e. stage II) accuracy considering all testing samples. We use \emph{stratified} five-fold cross validation to reduce selection bias and validate each model. In each fold, we randomly sample 80\% subjects from each class as the training set, and the remaining 20\% for each class as the test set. Within each fold, we employ 20\% of training data to optimize the hyper-parameters.

\section{Results}
\subsubsection{Quantitative comparison.}
We evaluate the effectiveness of the proposed self-supervised radiomics features on two classification tasks: a) discrimination of low grade and high grade of brain tumor and b) staging of lung cancer. 
\paragraph{Effectiveness of \emph{RE/SE} module.} From the first row of Table \ref{tab:results_SOTA}, one can observe that  traditional radiomics itself brings powerful features to quantify tumor characteristics.
On BraTS dataset, the comparison between traditional radiomics and vanilla self-supervised radiomics (\emph{3DSiam}) confirms our hypothesis that features learned by vanilla self-supervised method behave poorly, especially on the minor class (poor specificity). However, self-supervised radiomics with \emph{RE} or \emph{SE} module surpasses \emph{3DSiam} in specificity by a large margin.
The aggregation of the vanilla self-supervised representation and traditional radiomics does not show significant improvement. More importantly, with \emph{RE/SE} module added, the specificity increased by 6.6\%, from 64.5\% to 71.1\%, which indicates a large boost in predicting the minor class (i.e. L-grade). Both comparisons of rows 4,5,6 and rows 7,8,9 demonstrate the success of our RE/SE module in tacking class imbalance, i.e., promoting the recognition of the minor class, while preserving the accuracy of the major class. 
\paragraph{Comparison with state-of-the-art.} Our method (\emph{Trad.+3DSiam+SE} in Tab.~2) outperforms the supervised one in two scenarios 
in two classification tasks, the result of which is achieved by using the same encoder backbone with a weighted cross-entropy loss. 
When it is trained with 50\% less labels, the performance of supervised model decrease drastically. 
On lung cancer staging with three classes, although the overall accuracy of self-supervised radiomics is lower than the traditional one, with the \emph{RE/SE} module, the combination of two kinds of radiomics achieves the topmost overall accuracy. This demonstrates the proposed self-supervised radiomics is complementary to existing radiomics. In the second row, we show the result of one self-supervised learning method trained by playing Rubik cubes \cite{zhuang2019self} to learn contextual information with a same encoder. We observe that the representation learned in proxy task is less discriminative than the one directly from representation learning. 

\begin{table*}[t]
  \caption[table: Comparison]{\small Comparison of the performances of different kinds of features in two down-stream tasks using stratified cross-validation. We further reduce 50\% training data in each fold of validation to show the effectiveness against supervised learning. Our method outperforms supervised learning in both scenarios.}
  \label{tab:results_SOTA}
  \centering
  \setlength{\tabcolsep}{2mm}{
   \begin{tabular}{l | c c | c c}
    \hline 
     &\multicolumn{2}{c|}{\textit{BraTS}} & \multicolumn{2}{c}{\textit{Lung cancer staging}} \\
    \multirow{1}{*}{Methods} & \multicolumn{2}{c|}{Sensitivity/Specificity} & \multicolumn{2}{c}{Overall/Minor-class Accuracy} \\
   
     ~ & \textit{full labels} & \textit{50\% labels} & \textit{full labels} & \textit{50\% labels}           \\
    \hline
     Trad. radiomics & 0.888/0.697 & \textbf{0.848}/0.697 & 0.490/0.375& 0.481/0.325\\
     {Rubik's cube} \cite{zhuang2019self} &0.744/0.526 &0.680/0.486 & 0.459/0.325 & 0.433/0.275\\ 
     3DSiam & \cellcolor{red!10}0.844/0.407 & 0.808/0.526&\cellcolor{red!10}0.459/0.300 &0.445/0.300\\

     3DSiam+SE & \cellcolor{red!10}0.848/0.513 &0.824/0.566 &\cellcolor{red!10}0.471/0.350&0.443/0.325 \\
     3DSiam+RE & \cellcolor{red!10}0.868/0.486 & 0.828/0.605&\cellcolor{red!10}0.459/0.375& 0.445/0.325\\
     \hline
     Trad.+3DSiam & 0.904/0.645 & 0.804/0.566 & 0.495/0.350&0.486/0.350\\
     Trad.+3DSiam+SE & 0.916/\textbf{0.711} & \textbf{\cellcolor{blue!10}0.848/0.763} & \textbf{0.538}/0.375 &\textbf{\cellcolor{blue!10}0.519}/0.350\\
     Trad.+3DSiam+RE & \textbf{0.920}/\textbf{0.711} &0.804/0.763 & 0.524/\textbf{0.425}&0.502/\textbf{0.40}\\
     \hline
    Supervised Learning & {0.888/0.711}& 0.804/\cellcolor{blue!10}0.566 & 0.526/0.375 & \cellcolor{blue!10}0.467/0.325\\
     
     \hline
  \end{tabular}
}
\end{table*}



\subsubsection{Analysis of representations and hyperparameters.} 
\label{section_analysis}
\begin{figure}[t]
	\begin{center}
		\includegraphics[width=0.92\textwidth,height=0.620\textwidth]{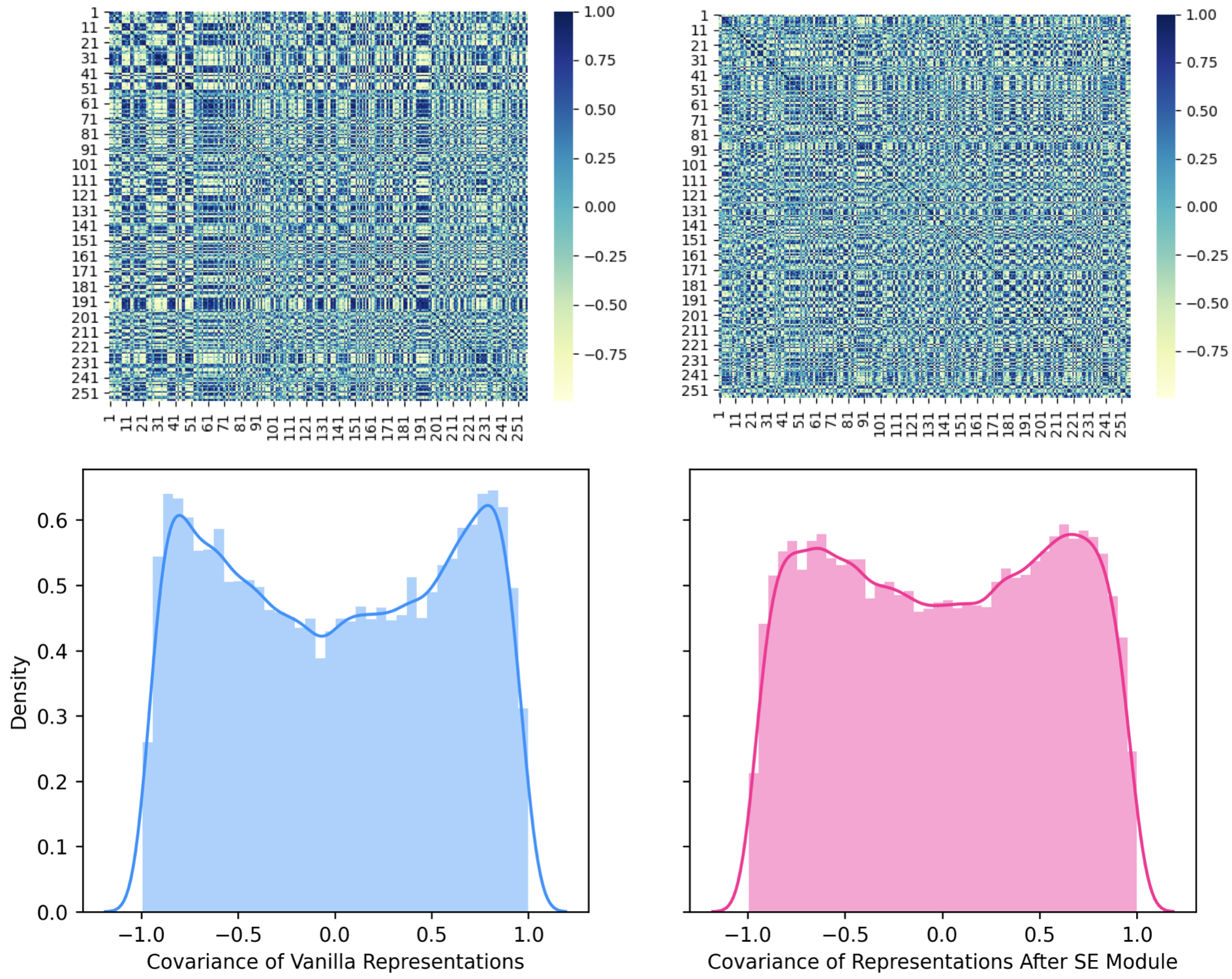}
	\end{center}
    	\caption{\small Covariance analysis of the representations before and after the \emph{SE} module. 
    	Across all 326 tumor patients, each feature was correlated with other ones, thereby generating the correlation coefficients. The density map show that the vanilla representation before \emph{SE} module are more correlated (redundant) than the one after.}
	\label{fig:analysis_Covariance} 
	\vspace{-0.5cm}
\end{figure}

\paragraph{Feature covariance.} For a better understanding of the role of the proposed module in relieving data imbalance problem, we further analyze the feature covariance to understand the role of the \emph{SE module}. 
Consider two paired variables $(x_{i},x_{j})$ in the representation $R$. 
Given $n$ samples $\{(x_{i1},x_{j1}), (x_{i2},x_{j2}), ..., (x_{in},x_{jn})\}$, Pearson's correlation coefficient ${r_{x_{i}x_{j}}}$ is defined as: ${r_{x_{i}x_{j}}}=\frac{cov(x_{i},x_{j})}{\sigma_{x_{i}}\sigma_{x_{j}}}$,
where $cov$ is the covariance and $\sigma$ is the standard deviation. 
We found that the features after \emph{SE} module become more compact as shown in Figure \ref{fig:analysis_Covariance} and more discriminative compared to the features without \emph{SE} module. 
\paragraph{Effect of the number of clusters k.} The hyper-parameter $k$ in the \emph{SE} module is the number of clusters, which plays a vital role in constructing new batch. To evaluate its effect, we use different $k$ to train \emph{3DSiam} and evaluate it through classification task. To fairly compare different values of $k$, we keep the size $m$ of the new batch ${B_c}$ fixed to $6$ which is also the batch size when $k=0$ (without \emph{SE} module). The initial batch size $N$ is set to k $\times$ q where q is empirically set to 10 in the comparison. The AUC achieves the highest when $k = 3$. With $k = 5$, the AUC drops. This is probably because when $k$ becomes large, the sampling may be biased when only considering a pair of clustering centers. For details, please refer to the curves of AUC over the number of clusters in Table 3 in Supplementary. 


\section{Conclusion}
In this work, we proposed a 3D \emph{self-supervised} representation framework for medical image analysis.
It allows us to learn effective 3D representations in a self-supervised manner while considering the imbalanced nature of medical datasets. We have demonstrated that data-driven self-supervised representation could enhance the predictive power of radiomics learned from \emph{large-scale} datasets without annotations and could serve as an effective compliment to the existing radiomics features for medical image analysis. 
Dealing with imbalance is an important topic and we will explore other strategies in the future. 

\section*{Acknowledgement}
This work was supported by Helmut Horten Foundation. I. E. was supported by the TRABIT network under the EU Marie Sklodowska-Curie program (Grant ID: 765148). S. L. was supported by the Faculty Research Grant (NO. FRG-18-020-FI) at Macau University of Science and Technology. B. W. and B. M. were supported through the DFG, SFB-824, subproject B12. K. C. was supported by Clinical Research Priority Program (CRPP) Grant on Artificial Intelligence in Oncological Imaging Network, University of Zurich.

\bibliographystyle{splncs04}
\bibliography{egbib}

\newpage



%
%
\section*{Supplementary Materials}

\begin{table*}[htpb]
  \caption[table: Comparison]{\small Class distribution of the BraTS dataset including patients with brain tumor. Not only the labels for the target task (i.e. 3:1), but also the subjects distribution among different centers, are imbalanced.}
  \label{tab:dataset}
  \centering
   \begin{tabular}{c c c c c c }
    \hline 
 Labels/Centers& ~CBICA~ & ~TCIA~ & ~~TMC~~ & others & ~{Total}~\\
    \hline
     H-grade  & 125 & 98 & 8 & 19 & 250\\
     L-grade  &  None & 65 & 1 & 10 & 76  \\
     \hline
  \end{tabular}
\end{table*}

\begin{figure}[]
	\begin{center}
		\includegraphics[width=1\textwidth,height=0.4\textwidth]{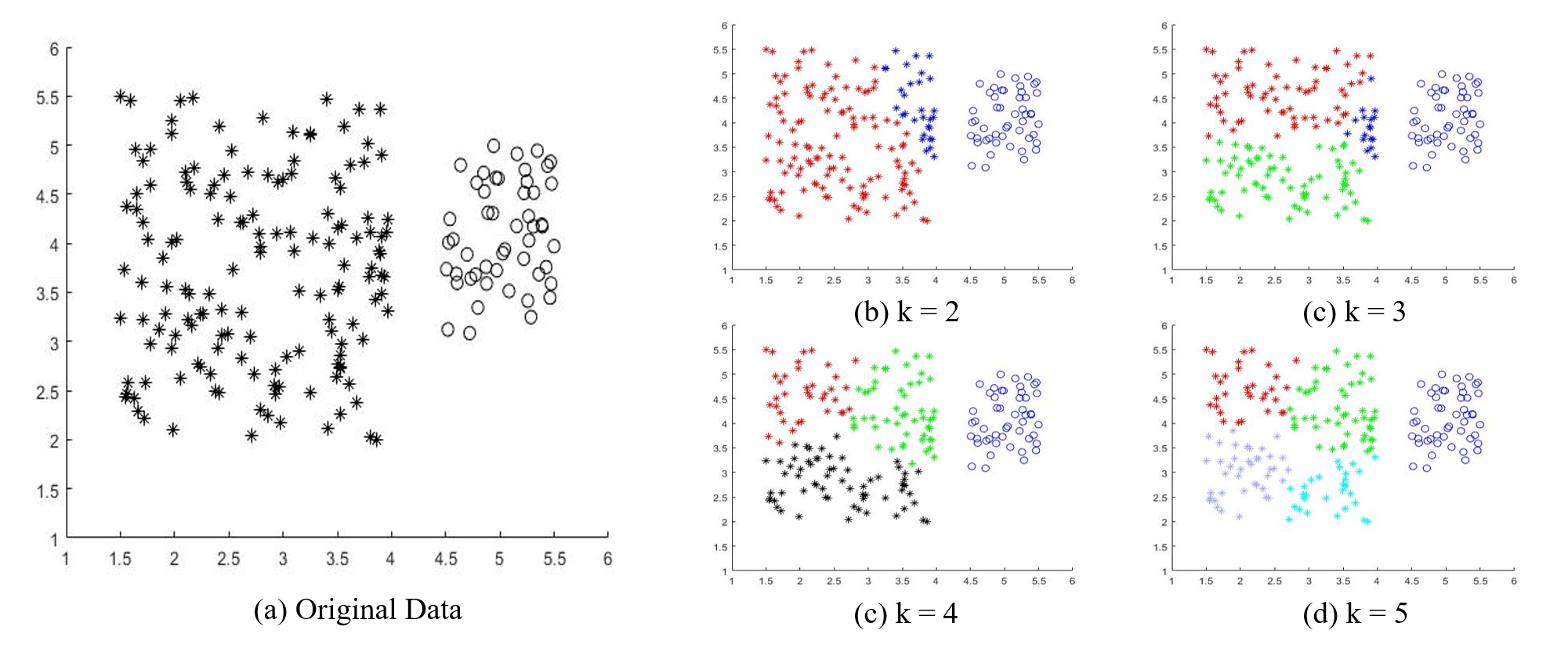}
	\end{center}
	\vspace{-0.3cm}
    	\caption{The illustration of clustering results of different values of $k$. Sub-figure (a) shows the class-imbalance data of a two-classes problem. Sub-figures (b) to (c) show the results of $k$-means with various values of $k$. Dots with different shapes belong to different classes. Different clusters are denoted by different colors. The centroids of the red and blue clusters are the farthest away, and the samples around them are selected to form the class-balanced input.}
	\label{fig:kmeans_results} 
\end{figure}
\vspace{-0.2cm}

\subsection*{Motivation behind \emph{SE} module}
When it comes to self-supervised learning, due to the random sampling strategy, models have much less chance to learn representations from the minor class than the major class. 
Existing state-of-the-art frameworks including \emph{SimSiam} \cite{chen2020exploring} are built upon balanced datasets without considering the class imbalance problem. 
To balance the training data in one batch, we propose a feature clustering based selection method. 

Without loss of generality, we illustrate our solution considering the binary classification problem. 
A set of class-imbalance data examples with underlying distributions is plotted as Fig. \ref{fig:kmeans_results} (a).  
The dots are clustered using $k$-means with different values of $k$, and the clustering results are shown in Fig. \ref{fig:kmeans_results} (b) to (e).
Intuitively, the clusters (red and blue) with the largest center distance would most probably consist of samples of different classes.
Therefore, a \emph{same} number of samples around the red and blue centers are selected to form a class-balanced training batch for the input to the self-supervised learning framework.
As shown in Table \ref{tab:label_analysis}, we calculate the real label distribution after sampling 2000 batches from the BraTS datasets. Batch size is set to different according to Section \ref{section_analysis}. We observed that the \emph{SE} module ease the imbalance issue (from 1:3.29 to 1:2.19) on BraTS dataset.

\begin{table*}[htpb]
  \caption[table:Analysis]{\small Analysis of real label distribution after SE module on BraTS dataset. The original data distribution of L-Grade and H-Grade is 1:3.29.}
  \label{tab:label_analysis}
  \centering
   \begin{tabular}{c | c c c c }
    \hline 
 k& ~2~ & ~3~ & ~~4~~ & 5\\
    \hline
     ratio (L:H)  & ~~1:3.34~~ & ~~1:2.19~~ & ~~1:2.33~~ & ~~1:2~~ \\
     \hline
  \end{tabular}
\end{table*}




\section*{Architecture}
\vspace{-0.2cm}

\begin{table}[H]
\centering
\setlength{\tabcolsep}{5mm}{
\begin{tabular}{ c|c|c} 
\multicolumn{2}{c|}{\textbf{Layers}} & \textbf{Output size} \\
\hline
$\text{conv}_1$& $3 \times 3 \times 3$, 32, stride 1,& $96 \times 96 \times 96$ \\ 
\hline
$\text{residual block}_2$ & $\left[\begin{array}{c}1 \times 1 \times 1,32 \\ 3 \times 3 \times 3,32 \\ 1 \times 1 \times 1, 32\end{array}\right] \times 1 $ & $96 \times 96 \times 96$ \\ 
\hline
$\text{residual block}_3$ & $\left[\begin{array}{c}1 \times 1 \times 1,32 \\ 3 \times 3 \times 3,32 \\ 1 \times 1 \times 1, 32\end{array}\right] \times 1 $   & $96 \times 96 \times 96$ \\ 
\hline
$\text{max-pooling}_3$ & $3 \times 3 \times 3$, stride 0 & $32 \times 32 \times 32$ \\ 
\hline
$\text{conv}_4$ & $3 \times 3 \times 3$, 64, stride 1, bn, relu  & $32 \times 32 \times 32$ \\ 
\hline
$\text{max-pooling}_4$ & $3 \times 3 \times 3$, stride 0 & $10 \times 10 \times 10$ \\ 
\hline
$\text{conv}_5$ & $3 \times 3 \times 3$, 128, stride 1, bn, relu  & $10 \times 10 \times 10$ \\ 
\hline
$\text{max-pooling}_5$ & $3 \times 3 \times 3$ , stride 0 & $3 \times 3 \times 3$ \\ 
\hline
$\text{conv}_6$ & $3 \times 3 \times 3$, 256, stride 1, bn, relu  & $3 \times 3 \times 3$ \\ 
\hline
\multicolumn{2}{c|}{global average pooling 3d} & $1 \times 1 \times 1$ \\
\hline
$\text{dense}_7$ & 324, bn & 324 \\ 
\hline
$\text{dense}_8$ & 256 & 256 \\
\hline
\multicolumn{3}{c}{}
\end{tabular}}
\caption{Our 3D encoder network for self-supervised representation learning. The dimensions of 3D output maps and filter kernels are in $T\times H\times W$, with the number of feature channels following. Residual blocks are shown in brackets. For demonstration, the input is set to $96\times 96\times 96$ but it can be arbitrary size in practice. \emph{conv} = convolutional layer, \emph{bn} = batch normalization layer. }
\label{table:backbone}
\end{table}

\section*{Analysis on Hyperparameters}
\vspace{-0.2cm}

\textit{Effect of batch size.} In addition to the effect of the number of cluster presented in the main manuscript, we evaluate the effect of different batch size $m$ of the selected batch by \emph{SE} module with the number of clusters $k$ set to 3. The AUC score changes slightly with different $m$. It is reasonable as the new formed batch still has more samples from the major class. When $m=2$, \emph{3DSiam} with SE module achieves the lowest AUC. With bigger $m$, the AUC increases and shows stable. It indicates that our model can be trained with small batch size within computation capacity. 
\vspace{-0.5cm}
\begin{figure}[H]
	\centering
	\includegraphics[width=110mm, scale=0.8]{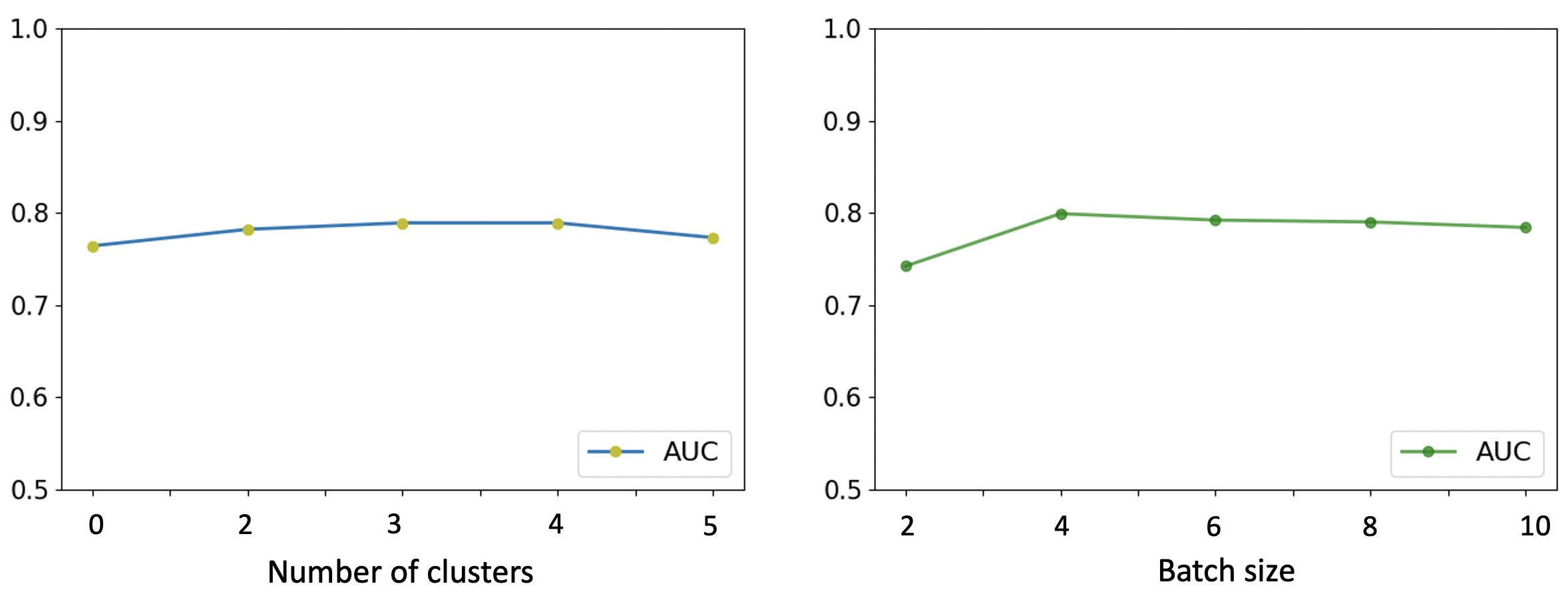}
	\vspace{-0.4cm}
    \caption{AUC score on BraTS dataset with different number of clusters or batch size. \emph{Left}: The effect of the number of clusters in SE. 0 cluster means the model is trained without SE module. \emph{Right}: The effect of batch size in the SE module. We observe that our approach is not sensitive to batch size when the size is larger than 4.}
	\label{fig:analysis_n_b} 
\end{figure}
\vspace{-0.2cm}

\section*{Additional Result}

\vspace{-0.2cm}
\begin{figure}[H]
	\centering
	\includegraphics[width=70mm,scale=0.8]{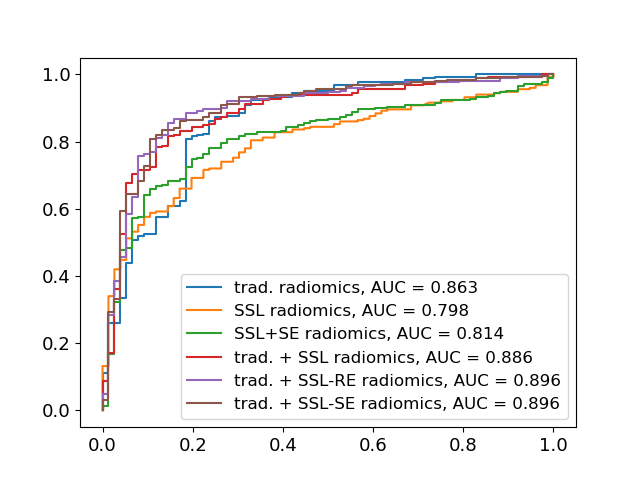}
    \caption{AUC scores on BraTS dataset achieved by different methods.}
	\label{fig:AUC_BraTS} 
\end{figure}




%
%

\end{document}